# Composable Core-sets for Diversity Approximation on Multi-Dataset Streams

arXiv Summarized Report: Spring 2023

**Stephanie Wang**
Applied Mathematics & Physics
University of Rochester
Rochester, NY
swang157@u.rochester.edu

**Michael Flynn**
Computer Science
University of Rochester
Rochester, NY
mflynn10@u.rochester.edu

**Fangyu Luo**
Computer Science
University of Rochester
Rochester, NY
fluo4@u.rochester.edu

## ABSTRACT

Core-sets refer to subsets of data that maximize some function that is commonly a diversity or group requirement. These subsets are used in place of the original data to accomplish a given task with comparable or even enhanced performance if biases are removed. Composable core-sets are core-sets with the property that subsets of the core set can be unioned together to obtain an approximation for the original data; lending themselves to be used for streamed or distributed data. Recent work has focused on the use of core-sets for training machine learning models. Preceding solutions such as CRAIG have been proven to approximate gradient descent while providing a reduced training time. In this paper, we introduce a core-set construction algorithm for constructing composable core-sets to summarize streamed data for use in active learning environments. If combined with techniques such as CRAIG and heuristics to enhance construction speed, composable core-sets could be used for real time training of models when the amount of sensor data is large. We provide empirical analysis by considering extrapolated data for the runtime of such a brute force algorithm. This algorithm is then analyzed for efficiency through averaged empirical regression and key results and improvements are suggested for further research on the topic.

## 1. INTRODUCTION

### 1.1 The State of Composable Core-sets

The era of big data has brought about an unprecedented increase in the volume, variety, and speed of data being generated and collected. This overflow of information has outpaced the capacity of traditional data processing and analysis techniques, necessitating the development of innovative methods to handle massive datasets efficiently. Among the various approaches proposed to tackle this issue, data summarization techniques have emerged as a promising solution. These techniques aim to reduce the size of the dataset while preserving its essential characteristics, enabling faster processing, and minimizing computational overhead. One of these techniques is called a "core-set," a term originally coined in 2005 by Agarwal, Har-Peled, and Varadarajan [1]. A core-set is a small, representative subset of a larger dataset that captures the properties and structure of the original data, such as uniformity and distribution. By constructing a core-set, we are able to reduce the computational complexity and processing time for various data-driven tasks, such as clustering, optimization, and machine learning, while maintaining the overall quality of the results.

Composable core-sets are an extension of the core-set concept, designed to offer additional advantages in terms of scalability and adaptability for parallel and distributed computing environments. The key difference between composable core-sets and traditional core-sets lies in their ability to be combined in a hierarchical manner while preserving their representative power. This leads to several advantages over traditional core-sets in terms of scalability, adaptability, and flexibility, which enables them to process large-scale and dynamic datasets with greater computational efficiency. However, both types of core-sets remain an approximation of their full dataset, and as such, contain an approximation error. The tradeoff between efficiency, size, and accuracy of representation then lies in the various algorithms and techniques used to construct the core-set in context of the problem it attempts to solve.

In this paper, we attempt effective representation of multiple streams through a single algorithm for the purpose of diversity-aware summarization. Currently, composable core-sets have been theorized to be effective at solving the diversity and coverage maximization problem [2]. Indyk, Mahabli, Mahdian, and Mirrokni's paper presents "off-line" algorithms GMM (Gaussian Mixture Model), Local Search, and Prefix that satisfy several diversity objectives; however, their approach has not been applied to real-world datasets and their results are proven computationally rather than empirically. This is likewise the result of Mirzasoleiman, Bilmes, and Leskovec's paper *Coresets for Data-efficient Training of Machine Learning Models* in where they were able to develop a method to select a weighted core-set that closely approximates the full gradient-descent vector but with no applications [3]. Moreover, experimental results that exist on core-set construction using existing, non-synthetic datasets have not been composable or are not purposed towards diversity [3][4][5]. The state of composable core-set research is decidedly sparse, thus we propose an algorithm that creates a composable core-set that approximates the diversity of the union of multiple, well-known CIFAR dataset as streams to additionally populate this area of research.

### 1.2 Our Contributions

- **Stream summarization using core-sets:** We consider the use of a reduced size adjacency matrix for establishing non-estimated results for solving the remote edge diversity problem in composable core sets. Composable core-sets lend themselves naturally to stream summarization. In this paper, we contribute two brute force algorithms for summarizing streams.

- **Implementation of core-set algorithms:** A common issue associated with core-sets is that they exist mostly in theoretical computer science. The algorithms discussed in this paper were implemented using python and were run on real data (i.e learned vector representations of the CIFAR10 dataset) in a synthetic testing environment.

- **Empirical analysis:** We consider extrapolated data to estimate how long such algorithms take to construct core-sets of size-k. The goal is to provide motivation for the exploration of heuristic and approximate solutions that can be used to train models more fairly and efficiently when data is in the form of many streams.

## 2. PRELIMINARIES

### 2.1 Problem Definition

Core-sets have varying and unfixed definitions within Computer Science. In Machine Learning, a core-set is defined as the minimal set of training samples that allows a supervised algorithm to deliver a result as accurate to its purpose as the one obtained when the whole set is used. Put simply, core-sets are subsets of data points that are representative of the original data set. Core sets may offer a promising solution to many issues associated with machine learning tasks such as long training times and skewed data[1]. Preceding work has focused on efficient core set construction algorithms for maintaining diversity requirements [1]. Separate work has shown that core sets can be used to speed up incremental gradient descent training by selecting core sets in such a way that they approximate the gradient of the original data [6]. In this paper, we will explore the combination of these two fields, to discuss core set algorithms that maintain diversity requirements as well as can be used for efficient training of machine learning models. The purpose will be to train models using streamed data, where the streams come from different data sources.

### 2.2 $\alpha$-Approximate Composability

Core-set composability is a property of certain core-set construction techniques that allows multiple core-sets to be efficiently combined into a new core-set that accurately represents the union of the underlying datasets. This property was first introduced by Agarwal et al. [1] and has since been studied and extended by various researchers in the field of data summarization and approximation algorithms [6]. Formally, given core-sets $\{C_1, C_2, ..., C_n\}$, each representing the datasets $\{D_1, D_2, ..., D_n\}$ respectively, core-set composability is satisfied if there exists an efficient algorithm to construct a new core-set C, such that C is a valid core-set for the combined dataset $\{D_1 \cup D_2 \cup ... \cup D_n\}$. Moreover, the size of C should be proportional to the size of each of the elements $\{C_1, C_2, ..., C_n\}$, and the approximation quality of C should be comparable to that of each element $\{C_1, C_2, ..., C_n\}$ with respect to a given problem or task. This composability property enables efficient parallel and distributed processing of large-scale datasets and has been successfully applied to a wide range of data-driven tasks, such as clustering [1], geometric approximation [6], and optimization [7]. In this paper, we specifically study $\alpha$-approximate composable core-set for a diversity objective. This is defined as a mapping from a set D to a subset of D with the following property: for a collection of sets, the maximum diversity of the $\cup$D is within an $\alpha$ factor of the maximum diversity of the union of its corresponding core-sets, $\cup$C.

### 2.3 Diversity Requirement: Remote Edge

Diversity requirements in data science refer to the need for diverse datasets, algorithms, and practitioners to ensure fairness, inclusivity, and unbiased outcomes in data-driven decision-making. The Remote-Edge problem is a criterion in graph-based data summarization and analysis that aims to ensure a diverse set of edges is selected, such that the selected edges span different parts of the graph while maintaining a significant distance from each other. This requirement is particularly relevant where incorporating diverse connections or relationships can lead to better performance and more robust solutions. It was first introduced by Mestre [8] in the context of the "uncapacitated facility location problem" and has since been studied and extended by various researchers in the field of network design and approximation algorithms [9][10][11]. While the Remote-Edge problem is not inherently a diversity requirement, it can support it by enabling diverse data sources to be processed and analyzed locally, closer to its origin. Furthermore, there has been a precedent in Indyk et al. 's paper where they considered Remote- Edge as a measurement of the diversity of composable core-set [4]. We define the Remote-Edge problem using Indyk et al.'s notion, $min_{p,q \in S} \; dist(p, q)$, and formalize it as follows to propound our definition of diversity.

***Definition 1.*** For any given set S and its subset S' ⊂ S, the diversity of S' is defined as $div(S') = \min_{p',q' \in S'} dist(p', q')$, where p', q' indicate points selected to be in the core-set, C. This diversity requirement ensures the following two qualities:

- $\min_{p',q' \in S'} dist(p', q') > \min_{p' \in S', \; q \in S \setminus S'} dist(p', q)$, or explained: the minimum distance between two points within the core-set C is not smaller than the minimum distance between a point ∈ C and a point ∉ C.
- $\min_{p',q' \in S'} dist(p', q') > \min_{p \in S \setminus S', \; q \in S \setminus S'} dist(p, q)$ or explained: the minimum distance between two points (p', q') ∈ C is not smaller than the minimum distance between two points (p, q) ∉ C.

### 2.4 Related Work

In *Composable Core-sets for Diversity and Coverage Maximization*, Indyk et al. present a Gaussian Mixture Model that does not use the standard Expectation-Maximization (EM) approach. Instead, it is a greedy algorithm that aims to find a subset of size k from the input set 'S' with the most diverse points, therefore ensuring diversity in the chosen subset [4]. They then go on to prove that their GMM algorithm satisfies the composability property. While initially Indyk et al.'s algorithm appears to present a solution to our application, it is not possible to directly implement GMM because it relies on the size of the dataset to be both known and fixed. However, through our work, we process the inputs as streams which are theoretically continuous, thereby establishing the need to present our unique algorithm.

## 3 FRAMEWORK INSTALLATIONS

### 3.1 Dataset and Streams

Our work makes use of the well-known CIFAR (Canadian Institute For Advanced Research) dataset, which is a collection of images used for machine learning and computer vision research [8][12]. It was created by researchers from the University of Toronto and consists of two subsets: CIFAR-10 and CIFAR-100. These datasets have become popular for benchmarking and testing machine learning algorithms, particularly in the areas of image classification and convolutional neural networks. In particular, our algorithm is evaluated on CIFAR-10, the smaller of the two datasets that consists of 60,000 32x32-pixel color images, divided into 10 classes. Each class represents a distinct object category, such as airplanes, automobiles, birds, cats, deer, dogs, frogs, horses, ships, and trucks. The dataset is split into a training set containing 50,000 images and a test set containing 10,000 random images from all classes. In each 32x32 pixel color image, each pixel has three color channels: red, green, and blue (RGB). In order to use these images as input for machine learning algorithms, the pixel values must be converted into a vector representation. This has already been done by Professor Nargesian's data selection group at the University of Rochester [13]. They were able to transform the CIFAR-10 image pixel values into vector representations using img2vec and the efficientnet_b0 training model, and we were able to obtain that corresponding 50,000 vector-dataset for our use. We further split the vector set into fifths where we treat each partition as individual datasets $\{D_1, D_2, D_3, D_4, D_5\}$, and disregard the test set as we did not use it to evaluate our algorithm. We further converted the datasets into simultaneous streams by feeding each element into the algorithm and assigning it a monotonically increasing time-stamp as it is processed.

### 3.2 Composable Core-set Construction

In the context of multiple input streams, the core-set construction process involves creating a core-set for each stream and then combining these core-sets at the end. This approach allows for the processing of each stream independently, leading to parallelism and improved efficiency. The size of the final core-set will be *k\*n*, where *k* is the size of each core-set and *n* is the number of input streams. Below, we introduce our **"Brute Force *k*-replacement"** and **"*k*-Adjacency Replacement"** algorithms which construct a composable core-set from streams by finding the **least satisfying points(s)** in the core-set to be replaced. To select a new point for the core-set once the core-set is full and find the least satisfying point(s) for it to replace, we require the following.

***Definition 2.*** For the two points *p'* and *q'* that form $\min_{p',q' \in C} dist(C)$, the minimum pairwise distance between any two points in the core-set *C*, the **least satisfying point** is *p'* or *q'* that has the $\min_{p',q',r' \in C} dist(p' \vee q', r')$ where *r'* is the new point selected to be in *C*. If dist(*p'*, *r'*) = dist(*q'*, *r'*), then *r'* will replace either *p'* or *q'* at random.

***Definition 3.*** Once the core-set *C* is full, to determine if a new point *r'* from the stream *S* is replacing**,** it must have the **eligibility**: For the two points *p'* and *q'* that form $\min_{p',q' \in C} dist(C)$, *r'* will replace *p'* or *q'* if [dist(*p'*, *r'*) ∨ dist(*q'*, *r'*)] > $\min_{r',s' \in C}$ dist(*r'*, *s'*), the minimum distance from *r'* to any one point in *C*. If accepted, *r'* will then replace *p'* if dist(*p'*, *r'*) < dist(*q'*, *r'*) and *q'* if dist(*p'*, *r'*) > dist(*q'*, *r'*). If dist(*p'*, *r'*) = dist(*q'*, *r'*), then *r'* will replace either *p'* or *q'* at random.

**Figure 1: k/k-Adjacency replacement algorithms' core-set selection and replacement process represented as graphs.**

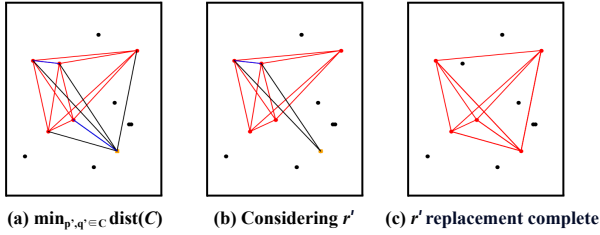

(a) $\min_{p',q' \in C}$ dist(*C*)   (b) Considering *r'*   (c) *r'* replacement complete

We present the above graphs to illustrate our algorithm's core-set selection and replacement methodology. **(a)** highlights the edges that represent $\min_{p',q' \in C}$ dist(*C*) and $\min_{r',s' \in C}$ dist(*r'*, *s'*) in blue, where *r'* is a yellow node. It is clear that $\min_{r',s' \in C}$ dist(*r'*, *s'*) > $\min_{p',q' \in C}$ dist(*C*), therefore *r'* will be accepted. In **(b)**, dist(*p'*, *r'*) and dist(*q'*, *r'*) are highlighted in green and it is clear the edge on the left is greater than the edge of the right. Consequently, in **(c)**, we see that the point that corresponds with the lesser edge has been removed from the core-set. The edges of the graph are calculated cosine similarities between points[2].

### 3.3 Brute Force *k*-replacement Algorithm

---
**Algorithm 1** Brute Force (k-replacement)
---
1: **Input** $S$: a set of streams = {$S_1, S_2, \ldots, S_i$}, $k$: size of the subset, $t$: time horizon
2: **Output** $C$: the union of $S_i$ subsets each of size $k$
3: **while** current_time < t **do**
4:   **for** each $S_i$ in $S$ **do**
5:     $s = S_i$.next()
6:     **if** $|C_i| < k$ **then**
7:       $C_i \leftarrow C_i \cup \{s\}$
8:     **else**
9:       Find $\min_E$ (the smallest edge in $E$), where $E = \{e_1, e_2, \ldots, e_i | (p', q' \in C_i)[e_i = \{p', q'\}]\}$.
10:      Find $\min_{E'}$ (the smallest edge of $E'$), where $E' = \{e'_1, e'_2, \ldots, e'_i | (p' \in C_i)[e'_i = \{p', s\}]\}$.
11:      **if** $\min_{E'} > \min_E$ **then**
12:        $e_x = \{p' \in \min_E, s\}$
13:        $e_y = \{q' \in \min_E, s\}$
14:        **if** $e_x < e_y$ **then**
15:          Replace $p' \in \min_E$ with $s$
16:        **else**
17:          Replace $q' \in \min_E$ with $s$
18:        **end if**
19:      **end if**
20:    **end if**
21:  **end for**
22: **end while**
23: **return** $C = C_1 \cup C_2 \cup \cdots \cup C_i$
---

[2] Further explanation of cosine similarity and edge representation will be expanded upon in **Section 4 Results**.

The **"Brute Force *k*-replacement"** algorithm below operates within a specified time horizon *t* and works on a set of input streams *S* and is designed to output a union of subsets *C*, where each subset $C_i$ consists of *k* points from the corresponding input stream $S_i$. The purpose of the algorithm is to maintain diversity in the subsets by minimizing the distances between the points within the subsets. The algorithm accomplishes this by following *Definition 1* and *Definition 2*. However, due to its brute force nature, the algorithm can be computationally expensive, especially when dealing with large datasets or high-dimensional data points.

### 3.4 *k*-Adjacency Replacement Algorithm[3]

*k*-**Adjacency Replacement** adheres to the same methodology as its Brute Force counterpart but adds an additional component of sophistication through the adjacency matrix that offers a slight improvement to the algorithm in terms of running-time (at the cost of space complexity).

**Figure 2(a): Adjacency Matrix Construction**

$$\begin{array}{c} & \begin{array}{ccc} A & B & C \end{array} \\ \begin{array}{c} A \\ B \\ C \end{array} & \left[ \begin{array}{ccc} sim(A,A) & sim(A,B) & sim(A,C) \\ sim(B,A) & sim(B,B) & sim(B,C) \\ sim(C,A) & sim((C,B) & sim(C,C) \end{array} \right] \end{array}$$

$$\downarrow$$

$$\begin{array}{c} & \begin{array}{ccc} A & B & C \end{array} \\ \begin{array}{c} A \\ B \\ C \end{array} & \left[ \begin{array}{ccc} 1 & Null & Null \\ sim(B,A) & 1 & Null \\ sim(C,A) & sim((C,B) & 1 \end{array} \right] \end{array}$$

As shown in **Figure 2(a)**, the adjacency matrix is constructed by taking the cosine similarity, the similarity of between two vectors of an inner product space, of two points. The similarity calculation is a combination and not permutation of 2-tuples, e.g. *sim(A,B) = sim(B,A)*. Furthermore, the similarity of a point to itself will always return a value of 1. Thus, it is evident that the adjacency matrix can be decomposed where only the lower triangular contains novel information while the main diagonal is a sequence of 1s and the upper triangular is symmetric to the lower. Note that *Null* in **Figure 2(a)** represents superfluous and unstored data, and does not indicate the use of null pointers in our program.

**Figure 2(b): Adjacency Matrix Construction Continued**

$$\begin{array}{c} \begin{array}{c} B \\ C \end{array} \left[ \begin{array}{cc} sim(B,A) & 1 \\ sim(C,A) & sim(C,B) \end{array} \right] \rightarrow \left[ \begin{array}{c} [sim(B,A)], \\ [sim(C,A), \quad sim(C,B)] \end{array} \right] \end{array}$$

[3] Implementation of the k-adjacency: https://github.com/mflynn840/Coreset-Construction.git

To eliminate unnecessary data storage, we then removed the first row and last column of the adjacency matrix. This still leaves some extraneous values in a visual, square matrix; however, within the program itself, those values would not be stored. Only the lower triangular of unique cosine similarity values exist within the adjacency matrix.

```
Algorithm 2  k-Adjacency Replacement
1:  Input S: a set of streams = {S_1, S_2, ..., S_i}, k: size of the subset, t: time
    horizon
2:  Output C: the union of S_i subsets each of size k
3:  while current_time < t do
4:      for each S_i in S do
5:          s ← S_i.next()
6:          if |C_i| < k then
7:              C_i ← C_i ∪ {s}
8:          else
9:              if adjacency matrix is None then
10:                 Create adjacency matrix A of C_i
11:             end if
12:             Find min_E = arg min A[i][j]
13:             Create adjacency column of s with rows being c_1, c_2, ..., c_i ∈ C_i
14:             Find min'_E = arg min adjacency column of s
15:             if min'_E > min_E then
16:                 e_x = {p' ∈ min_E, s}
17:             else
18:                 e_y = {q' ∈ min_E, s}
19:             end if
20:             if e_x < e_y then
21:                 Replace p' ∈ min_E with s
22:                 Replace all similarities in matrix containing p' with those con-
    taining s
23:             else
24:                 Replace q' ∈ min_E with s
25:                 Update column q' of A with the adjacency column of s
26:             end if
27:         end if
28:     end for
29: end while
30: return C = C_1 ∪ C_2 ∪ ... ∪ C_i
```

The **"$k$-Adjacency Replacement"** algorithm avoids computing pairwise cosine similarities at each iteration by keeping a record of the pairwise similarities of all points in the core-set in the minimal amount of storage space. When the minimum edge distance needs to be found, the problem can be reduced to finding the smallest element in a $(k^2 - k) / 2$ matrix As points are replaced in the core-set, the adjacency matrix is updated by computing only $k$ cosine similarities. However, constructing the matrix (which is only done a single time when the core-sets cardinality becomes $k$) is expensive so the algorithm's time savings pay off more as the algorithm continues to run. Finally, there are memory limitations with this approach when the size of the core-set is large.

## 4 RESULTS

Our primary results provide an empirically extrapolated estimation of the running time of the k-adjacency replacement algorithm. Unfortunately, past a certain size of $k$, the core-set construction algorithm did not produce results in a feasible amount of time. Therefore, data was collected for various (feasible) values of $k$, and the data was extrapolated to estimate the runtime of the algorithm for varying sizes of coresets. The data used to construct the core-sets were learned vector representations from the CIFAR-10 dataset. To simulate the streaming environment, the data was stored using a stream data structure that fed one element at a time into the algorithm, and masked the underlying full dataset. The CIFAR10 dataset is split into 5 batches, so naturally, each batch was represented as a stream. Each "stream" contained 10,000 vectors. The two primary costs associated with the k-adjacency algorithm are construction of the adjacency matrix which occurs once per stream. The algorithm was run on core-set sizes of 10, 50, 100, 500 and 1000 for 5 streams. The adjacency matrix construction time for each stream was averaged together for each of the sizes. A linear regression was run on this data to extrapolate a function for the average adjacency matrix construction time as a function of the core-set size in **Figure 3 (a)**.

**Figure 3 (a): Approximation of the average of adjacency matrix construction time to the size of the core-set**

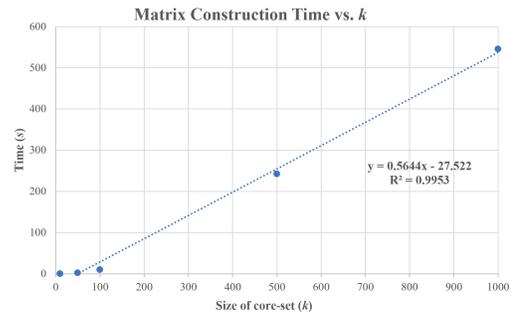

The second key step of the adjacency matrix core-set algorithm is updating the matrix and core-set when incoming points are found to be more beneficial than those in the core-set. In adjacency matrix core-set construction, this problem reduces to finding the smallest element in a $(k^2 - k) / 2$ matrix. Updating the matrix reduces to compute $k$ cosine similarities, and accessing the coordinates that need to be replaced can be done in constant time using dictionaries. The process of updating elements in the core-set (and by extension the adjacency matrix) was recorded for feasible values of $k$, and averaged together from each of the 5 streams to be extrapolated and get a function for the amount of time to replace a single element in the core-set as a function of $k$.

**Figure 3 (b): Approximation of the average core-set point selection and replacement time to the size of the core-set**

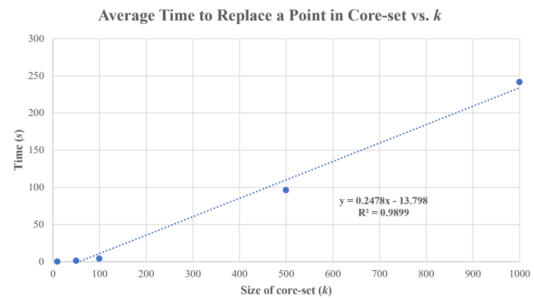

Using the two obtained functions previously described, an estimate for the total core-set construction time was extrapolated as a function of k, by assuming that 90% of elements that come in will require adjusting the core-set (and that this proportion remains fixed). The function is shown as follows along with its graph.

**Figure 3 (c): Approximation for total core-set construction time as a function of the size of the core-set ($k$)**

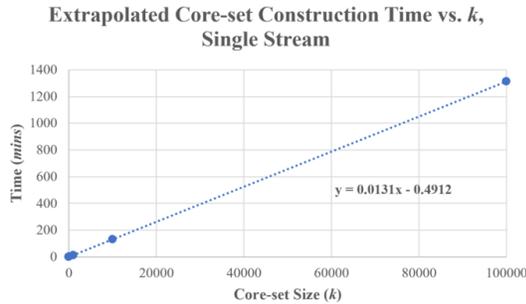

## 5 CONCLUSIONS AND FUTURE RESEARCH

With these results, we conclude that it is likely unfeasible to use brute force algorithms such as **Algorithm 1** and **Algorithm 2** to reduce the size of training sets. A typical CIFAR10 classifier takes approximately 8 hours to train, therefore a constraint on the core-set construction is that it must take less time than the savings in training time that it will produce. This is unlikely for the algorithms presented in this paper. Therefore, more efficient algorithms for summarizing streamed data should be explored in future works. Furthermore, bounding the error on model performance is a common issue with using core sets as training data. Therefore, techniques should be implemented such as those from CRAIG to approximate the gradient of the data for provable error bounds. Finally, faster algorithms as well as approximate solutions and heuristics should be explored for constructing composable core sets in a feasible amount of time.

### ACKNOWLEDGMENTS

The formation of this research problem and its subsequent result-measuring methodology was aided in great part by Professor Fatemeh Nargesian of the University of Rochester's Department of Computer Science with additional support provided by her data selection group.

### REFERENCES

[1] Dan Feldman. Introduction to core-sets: an updated survey. arXiv preprint arXiv:2011.09384, 2020.
[2] Indyk, P., Mahabadi, S., Mahdian, M., & Mirrokni, V. S. (2014). Composable core-sets for diversity and coverage maximization. In Proceedings of the 33rd acm sigmod-sigactsigart symposium on principles of database systems (pp. 100–108).
[3] Baharan Mirzasoleiman, Jeff Bilmes, and Jure Leskovec. Coresets for data-efficient training of machine learning models. In International Conference on Machine Learning, pp. 6950–6960. PMLR, 2020a.
[4] Indyk, P., Mahabadi, S., Oveis Gharan, S., and Rezaei, A. Composable Core-sets for Determinant Maximization: A Simple Near-Optimal Algorithm. In International Conference on Machine Learning (ICML), 2019.
[5] Barger, A., & Feldman, D. (2016). k-means for streaming and distributed big sparse data. In Proc. of the 2016 siam international conference on data mining (sdm'16).
[6] Har-Peled, S., & Mazumdar, S. (2004). On coresets for k-means and k-median clustering. Proceedings of the Thirty-Sixth Annual ACM Symposium on Theory of Computing (pp. 291-300).
[7] Feldman, D., Monemizadeh, M., Sohler, C., & Woodruff, D. P. (2011). Coresets and sketches for high dimensional subspace approximation problems. Proceedings of the Twenty-Second Annual ACM-SIAM Symposium on Discrete Algorithms (pp. 630-649).
[8] Mestre, J. (2006). Greedy in approximation algorithms. Proceedings of the 14th Annual European Symposium on Algorithms (ESA 2006) (pp. 528-539).
[9] Chudak, F., & Williamson, D. (2006). Improved approximation algorithms for the uncapacitated facility location problem. SIAM Journal on Computing, 35(4), 995-1008.
[10] Svitkina, Z., & Tardos, É. (2010). Facility location with hierarchical facility costs. ACM Transactions on Algorithms (TALG), 6(2), 1-21.
[11] Gupta, A., Krishnaswamy, R., & Molinaro, M. (2010). Approximation algorithms for correlated knapsacks and non-martingale bandits. Proceedings of the 51st Annual IEEE Symposium on Foundations of Computer Science (FOCS) (pp. 827-834).
[12] Krizhevsky, A., & Hinton, G. (n.d.). CIFAR-10 and CIFAR-100 datasets. University of Toronto. Retrieved from https://www.cs.toronto.edu/~kriz/cifar.html
[13] Safka, Christian. "Image 2 Vec with PyTorch." Commit 8b63ed6, img2vec, Dec 17, 2022, https://github.com/christiansafka/img2vec